\pdfoutput=1

\documentclass[11pt]{article}

\usepackage{EACL2023}
\usepackage{amsmath}
\usepackage{amssymb}
\usepackage{microtype}
\usepackage{esvect}
\usepackage{latexsym}
\usepackage{mathtools}
\usepackage{tcolorbox}
\usepackage{breqn}
\usepackage{graphicx}
\usepackage{algorithm}
\usepackage{algorithmic}
\usepackage{esvect}
\usepackage{latexsym}
\usepackage{mathtools}
\usepackage{tcolorbox}
\usepackage{breqn}
\usepackage{makecell}
\usepackage{multirow}
\usepackage{amsmath}
\usepackage{amssymb}
\usepackage{subcaption}
\usepackage{booktabs}
\usepackage{array}
\usepackage{threeparttable}

\usepackage{array}
\usepackage{placeins}
\usepackage{booktabs,arydshln}
\usepackage{float}
\newcolumntype{?}{!{\vrule width 1pt}}

\usepackage{times}
\usepackage{latexsym}

\usepackage[T1]{fontenc}

\usepackage[utf8]{inputenc}

\usepackage{microtype}

\title{FewShotTextGCN: K-hop neighborhood regularization for \\few-shot learning on graphs}

 \author{Niels van der Heijden$^{\clubsuit\diamondsuit}$  ~ Ekaterina Shutova$^{\clubsuit}$ ~  Helen Yannakoudakis$^{\spadesuit}$\\
$^\clubsuit$ILLC, University of Amsterdam, the Netherlands \\
$^\spadesuit$Dept. of Informatics, King's College London, United Kingdom \\
$^\diamondsuit$Deloitte Risk Advisory, Amsterdam, the Netherlands \\
{ \small \tt n.vanderheijden@uva.nl, e.shutova@uva.nl, } \\
{ \small \tt helen.yannakoudakis@kcl.ac.uk}
}

\begin{document}
\maketitle
\begin{abstract}
We present FewShotTextGCN, a novel method designed to effectively utilize the properties of word--document graphs for improved learning in low-resource settings. We introduce K-hop Neighborhood Regularization, a regularizer for heterogeneous graphs, and show that it stabilizes and improves learning when only a few training samples are available. We furthermore propose a simplification in the graph-construction method, which results in a graph that is $\sim$7 times less dense and yields better performance in low-resource settings while performing on-par with the state of the art in high-resource settings. Finally, we introduce a new variant of Adaptive Pseudo-Labeling tailored for word--document graphs.
When using as little as 20 samples for training, we outperform a strong TextGCN baseline with 17\% in absolute accuracy on average over eight languages. We demonstrate that our method can be applied to document classification without any language model pretraining on a wide range of typologically diverse languages while performing on par with large pretrained language models.
\end{abstract}

\section{Introduction}

Text classification, a key task in natural language processing (NLP), has many real-world applications, including toxic comment identification, news categorization, spam detection and opinion mining. One popular approach to this problem relies on large-scale pretraining of Transformer models 
\cite{devlin2018bert,conneau2019unsupervised,raffel2020exploring},  which have shown to be able to approach or even surpass human performance on many natural language understanding (NLU) benchmarks \citep{rajpurkar2016squad,wang2019superglue,liang2020xglue}. While these results are impressive for the languages on which models are pretrained, performance tends to deteriorate on languages where no or little data is available \citep{chau2021specializing,van2020comparison}. In practice, this means that these models are effective on a set of approximately 100 out of the 7000+ spoken languages in the world. Next to the requirement for vast amounts of data for pretraining, Transformer language models tend to be impractically large in terms of their number of parameters and have a high environmental footprint \cite{strubell-etal-2019-energy}.  

\noindent Recently, Graph Neural Networks (GNNs) have shown to be effective for text classification in both transductive \citep{yao2019graph,liu2020tensor,lin2021bertgcn} and inductive \citep{nikolentzos2020message,ding2020more} learning settings -- with promising results in both high- and low-resource settings. Particularly in the transductive setting, the authors of TextGCN \citep{yao2019graph} show that Graph Convolutional Networks (GCNs) \citep{kipf2016semi} can outperform state-of-the-art methods for document classification on English datasets without any language model pretraining. They do so by modeling an entire corpus of documents simultaneously as one heterogeneous word--document graph. The document classification task is formulated as a node-classification task over this graph. 

\noindent Later work shows that (multilingual) Pretrained Language Models (mPLMs) can be used to provide GNNs used in transductive setting with rich representations of both words and documents, improving results further in both monolingual \citep{lin2021bertgcn} and cross-lingual settings \citep{wang2021cross,li2020learn}. These works focus solely on high-resource settings and do not report any results on performance in low-resource settings. 

\noindent In this work, we propose a novel 
GNN-based method for learning document classification tasks on a range of languages without the need for any pretraining data (i.e., without utilizing any pretrained word embeddings or language models), and from few labeled samples only. To the best of our knowledge, we are the first to investigate few-shot graph-based transductive document classification in a range of languages other than English.

\noindent We present FewShotTextGCN, an improved version of the original TextGCN model, where we exploit properties of the heterogeneous word--document graph for improved learning from scratch and with only a few labels. More specifically, we: 

(1) Introduce K-hop Neighborhood Regularization (K-NR), an unsupervised learning technique for heterogeneous graphs, and use it in its $K=2$ instantiation as a regularizer tailored to word--document graphs , and show that it consistently leads to performance gains in low-resource settings;

(2) Propose a simplification of the graph-construction method, which results in improved performance in the low-resource setting while reducing the graph density by a factor of approximately 7 on average, therefore substantially speeding up computations and reducing memory requirements;

(3) Present a variant of adaptive pseudo-labeling \citep{zhou2019effective} on word--document graphs and show that it leads to consistent gains over the original TextGCN approach \citep{yao2019graph}, particularly when combined with K-NR. 

\noindent We compare FewShotTextGCN to its predecessor and two strong PLMs on ten topic classification benchmarks comprising eight typologically diverse languages, and experiment with a range of low-resource settings, including using as little as 20 labeled samples to learn from, and without any other form of (pre-trained) knowledge about a language except for what constitutes a word (using word boundaries or a tokenizer). 
In our lowest-resource setting, our method outperforms TextGCN with 4.6\% and 17\% points in absolute accuracy on average for Reuters and MLDoc, respectively -- while having a substantially smaller computational and memory footprint.
FewShotTextGCN performs on par with large PLMs on the great majority of the considered benchmarks, without the need for any large-scale pretraining and at only a fraction of the parameter count of these PLMs -- indicating that graph-based methods are an attractive alternative to using large PLMs for topic classification.
All our code and models are released to facilitate further research on this topic.\footnote{\url{https://github.com/mrvoh/FewShotTextGCN}}

\section{Graph Neural Networks}
Graph Neural Networks (GNNs) are a class of neural models designed to facilitate representation learning on geometric data -- data that naturally occur in many situations/fields, such as chemistry, social networks, maps, visual meshes, etc. Recently, there has been a great surge in research on GNNs. 
GNNs create new feature representations of nodes by aggregating the nodes' own feature representation and a message passed on from neighboring nodes. A graph $\mathcal{G} = (\mathcal{V}, \mathcal{E})$ is defined as a set of nodes $\mathcal{V}$ with edges $\mathcal{E}$ between them, typically represented as a square adjacency matrix A, where each entry holds the weight of the edge between node $j$ and $i$. Locality in graphs is defined by neighborhoods, where the neighbors of node $i$ are defined as $\mathcal{N}_i = \{j : (i, j) \in \mathcal{E} \,\, \forall (j, i) \in \mathcal{E} \}$. 
Let $\bigoplus$ be some permutation invariant aggregator such as $sum$, $average$ or $max$, and let $\psi$ and $\phi$ be two differentiable, learnable functions such as an MLP. 
Using these ingredients, we can describe GNNs by the way they do message passing.


\noindent Convolutional GNNs use the weights $c_{ij}$ of the edge between nodes $j$ and $i$ to weigh the incoming messages. These weights are part of the definition of the graph, meaning they are statically defined. The input feature $x_i$ of node $i$ is transformed to a latent representation $h_i$ by taking
\begin{align}
   h_i = \phi(x_i,\bigoplus_{j \in \mathcal{N}_i} c_{ij}\psi(x_j)) \label{eq:gcn}
\end{align}
The first and most well-known convolutional GNN is the Graph Convolutional Network (GCN) \citep{kipf2016semi}.

\section{Related work} \label{sec:rel work}
Our work is based on TextGCN \citep{yao2019graph}, which also serves as our baseline for all experiments. To the best of our knowledge, we are the first to investigate few-shot graph-based transductive learning from scratch for document classification in a range of languages other than English. Since the scope of our work is few-shot document classification in many languages by learning from scratch, we do not consider CLHG \citep{wang2021cross} directly related work. The reasons being it models corpora in multiple languages jointly, whereas we learn each task in isolation, and relies on machine translation and mPLMs. Similarly, MGL \citep{li2020learn} relies on mPLMs for encoding similar corpora in different languages into one embedding space, where consecutively a graph is dynamically constructed based on the similarity of the documents in the respective embedding space. Finally, meta-learning is applied to learn to classify documents in one language, given a limited set of documents in at least three other languages. Hence, we do not review these works in-depth.

\paragraph{TextGCN}
TextGCN \citep{yao2019graph} is the first application of GNNs to transductive text classification, applied on English datasets without any language model pretraining. The great majority of experiments is performed in high-resource settings, but a small set of results on performance in low-resource settings is also provided -- motivating us to further explore and expand upon this subject.
The authors construct a heterogeneous graph containing both word and document nodes. Word--word edges are weighed based on the pointwise mutual information (PMI) between the respective words, and word--document edges are created based on the TF--IDF score of the word in the respective document. More specifically, the adjacency matrix A is defined as:
\begin{align}
A_{ij} = 
\begin{cases}
\quad\text{PMI}(i, j)& i,j \text{ words, } \text{PMI}(i, j) > 0\\
\quad \text{TF--IDF}_{ij}& i \text{ document, } j \text{ word }\\
\quad 1 & i = j \\
\quad 0&  \text{otherwise}
\end{cases}
\end{align}
Document--document links are not considered. A one-hot encoding is used as input features for the nodes and a two-layer GCN is used to classify the document nodes. While this setup is relatively simple in terms of preprocessing, pretraining and the number of parameters in the model, the authors show that their method performs on par with state-of-the-art methods, even improving the state of the art for the 20News\footnote{\url{http://qwone.com/~jason/20Newsgroups/}} dataset. 

\paragraph{BERTGCN}
Follow-up work on TextGCN is that of BERTGCN \citep{lin2021bertgcn}, where the authors leverage PLMs to initialize document--node features. More specifically, a BERT-based model is used to encode the documents, and all other nodes are initialized with a one-hot vector. The BERT model used for encoding documents is optimized both via gradients propagated through the GCN and via an auxiliary classifier that directly uses the BERT embeddings to classify the documents. Using BERTGCN, the authors improve over TextGCN on a variety of text classification tasks -- especially on a sentiment analysis task, for which word order information is crucial for good performance \citep{johnson2014effective}. To be able to use BERTGCN in a full-batch gradient descent method, the authors use a memory bank that allows decoupling the dictionary size from the mini-batch size. 
Although the presented results are promising, a drawback of using large PLMs is the need for vast amounts of pretraining data, making these methods inaccessible for low-resource languages. 

\section{Data} \label{sec:data}
In this section, we give an overview of the datasets we use and the respective classification tasks.

\paragraph{MLDoc}
\citet{schwenk2018corpus} published an improved version of the Reuters Corpus Volume 2 \citep{lewis2004rcv1} with balanced class priors for all languages.  
MLDoc consists of news stories in 8 languages: 
English, Spanish, French, German, Italian, Russian, Japanese and Chinese. 
Each news story is manually classified into one of four classes: \textit{Corporate/Industrial (CCAT), Economics (ECAT), Government/Social (GCAT)} and \textit{Markets (MCAT)}. Per language, the train and test datasets contain 1k and 4k samples respectively. 

\paragraph{Reuters 21578}
From the Reuters-21578 dataset, a dataset of English news articles on a wide variety of topics, we use the \textbf{R8} and \textbf{R52} subsets (all-terms versions\footnote{\scriptsize{{\url{https://ana.cachopo.org/datasets-for-single-label-text-categorization}}}}). R8 has 8 categories and consists of 5485 and 2189 samples for training and testing respectively. R52 has 52 categories and 6532 and 2568 samples for training and testing respectively. The distribution of samples over the respective categories is highly skewed. \\
\vspace{-0.2cm}

\noindent During preprocessing on both datasets for all GNN-based models, we remove words with a frequency of less than 
5, 
and tokenize the data. For all languages except Japanese and Chinese, we split sentences based on whitespace. For Chinese, we use the Jieba\footnote{\url{https://github.com/fxsjy/jieba}} tokenizer, and for Japanese, the Fugashi one \citep{mccann2020fugashi}. For Transformer-based models, solely their respective tokenizers are used.


\section{Approach}
\paragraph{Graph construction} \label{sec:graph_construction}
We follow the graph construction method as described in the original TextGCN \citep{yao2019graph} work except we deviate in two different directions. Stopword removal is omitted, as this assumes knowledge of the language, whereas we aim for an approach that assumes no prior knowledge. Furthermore,  word--word edges are omitted too. Omitting such edges results in a much less densely connected graph, making learning substantially less memory intensive. We argue that the added value of word--word edges in a word--document graph is minimal given 1) only global information of word co-occurrence is considered (i.e., co-occurrence over the whole corpus as opposed to within document co-occurrence), and 2) over the course of training, words co-occurring in a document can influence each other's representation through an $N$-layer GNN where $N>1$. We illustrate this with an example in Appendix \ref{sec:appendix-word-word}, while in Section {\ref{sec:word-word-edges}}, we experimentally demonstrate the limited effect of word--word edges using ablation studies. 

\paragraph{K-hop Neighborhood Regularization (K-NR)}
We propose a new method that can exploit the properties of a word--document graph, inspired by approaches such as GraphSage \citep{hamilton2017inductive} that shows that meaningful node representations can be learned in an unsupervised manner with contrastive learning methods like Node2Vec \citep{grover2016node2vec}. These methods typically consist of two components: a sampling technique for deciding what nodes are regarded as positive or negative samples, and a loss function. Let $u$ be the anchor node, $P_p$ the positive sample sampling method, $P_n$ the negative sample sampling method, and $\mathcal{J}_{\mathcal{G}}(u, P_p, P_n)$ the contrastive loss function. In the case of GraphSage, $P_p$ is defined as a random walk starting from the anchor node, and $P_n$ is defined as uniformly sampling from all available nodes.

\noindent This contrastive learning approach on graphs assumes homogeneity and $P_p$ always samples in the 1-hop neighborhood from the anchor node. Herein, we propose a contrastive learning regularization method tailored on heterogeneous graphs instead, where nodes of the same type are K-hops away from each other on the graph. 

\noindent In what follows, we describe our approach in detail for heterogeneous word--documents graphs for the specific case of K-NR with $K=2$.
Driven by the intuition that documents (within a language) that share large parts of their vocabulary are more likely to be about the same topic, we introduce 2-hop Neighborhood Regularization (2-NR), a novel unsupervised learning method which can be used as a regularization technique. 

\noindent Let $\mathcal{G} = (\mathcal{V}, \mathcal{E})$ be the graph defined by the vertices $\mathcal{V}$ and edges $\mathcal{E}$. Let $\mathcal{V}_d, \mathcal{V}_w$ be the document and word nodes respectively. Given anchor node $u \in \mathcal{V}_d$, we first sample a word node $v \in \mathcal{V}_w$ connected to $u$ by sampling from a multinomial distribution weighted by the edge attribute values (the TF--IDF scores): 
\begin{align}
    v \sim Multinomial(1, A_{u,\{w| w \in \mathcal{V}_w \land w \in \mathcal{N}(v) \}})
\end{align}
Then, a positive document node $u_p$ and negative document node $u_n$ are sampled as follows:
\begin{align}
    u_p \sim U(\mathcal{N}(v)) \\
    u_n \sim U(\mathcal{V}_d \setminus \mathcal{N}(v))
\end{align}
Let $z_u$ be the final hidden representation of node $u$, the 2-NR loss, $\mathcal{L}_{2\textrm{-}NR}$, is then defined as:
\begin{multline}
    \mathcal{L}_{2\textrm{-}NR}(u, u_p, u_n) = \\max\{d(u, u_p) - d(u, u_n) + m, 0 \}
\end{multline}
\noindent for some distance function $d$ and margin $m$. This represents a triplet margin loss \citep{balntas2016learning}, which forces $u$ to be closer to $u_p$ than $u_n$ by at least a margin $m$.
See Appendix \ref{sec:appendix_2nr} for an elaboration on the intuition of 2-NR as well as a visualization.
In the word--document graph case, $K=2$ works specifically because we know that document nodes are only connected to word nodes and vice versa (see Section \ref{sec:graph_construction} for a description of our graph construction method). Hence, when starting at a document node, all nodes in its neighborhood are word nodes and similarly, all those word nodes do exclusively have edges to document nodes. Hence any walk of two steps starting at some document, will end up at another document via a word (node) they both contain. This simplifies the implementation for our specific word--document graph, but one can easily imagine generalizing the method to situations where taking $K$ hops on the graph does not guarantee ending up at a node of the same type as the start node by restricting the sampling methods to a subset of the desired nodes.

\paragraph{Adaptive pseudo-labeling}
Pseudo-labeling is a well-explored technique for improving performance in semi-supervised learning settings \citep{lee2013pseudo}, which, recently, has also been successfully applied to graphs \citep{zhou2019effective,chen2021graph}. We argue this technique can be particularly powerful for heterogeneous word--document graphs based on three premises:

(1) Different topics/classes have a different distribution of words in their vocabulary. 
    So it can be assumed that there exist words per class that occur more often in documents corresponding to that respective class -- i.e. these words are more distinctive for that given class, which in the word--document graph translates to that word node having relatively more edges to documents of the class the respective word is distinctive for.
    
(2) Document nodes are always at least two hops away from each other in the graph, meaning that only the input features of one document can influence the final feature representation of another document via message passing on the graph. This is assuming a two-layer GNN.

(3) The most effective way of encoding label information in the input document embedding is by directly optimizing for that respective class on the node, as opposed to relying on indirect optimization via backpropagating through the message-passing computational graph.

\noindent Instead of applying adaptive pseudo-labeling to the whole graph, we propose to only apply it to a subset of unlabeled document nodes, $\mathcal{U}_d$, that are not part of our train or test split. By doing this, we can directly optimize an unlabeled document embedding to be a good predictor for a certain class (premise (3)) .  This class-tailored document embedding can now be propagated over the graph to be used in the final feature representation of other document nodes via message passing on the graph (premise (2)). Finally, we can assume that there exist word nodes in the graph which are characteristic of a topic/class and via which the class-specific features can be propagated to other documents without losing information due to over-smoothing (premise (1)).

\noindent We implement adaptive pseudo-labeling as described by \citet{zhou2019effective}, which adds an extra component to the total loss, the pseudo-label loss $\mathcal{L}_{pse}$:
\begin{align}
    \mathcal{L}_{pse} = \sum_{v_i \in U^' } \frac{1}{N_i} CE(\tilde{Y}_i, F_i)
\end{align}
With $CE$ representing the cross-entropy loss, $\tilde{Y}_i$ the pseudo-label and $F_i  \in \mathbb{R}^C$ the predicted probability per class. The pseudo-label is generated by taking the argmax over $F_i$, which results in the pseudo-label loss optimizing for high-confidence predictions on the most certain class. The set of unlabeled samples $U'$ used for this loss is defined as:
\begin{align}
    U^'= \{u_i: u_i \in U_d | F_{i,j} \leq \beta \}, j = \underset{j^'}{argmax}F_{i, j^'} \}
\end{align}
Some minimum confidence threshold $\beta$ is used to filter out predictions, and the pseudo-loss per node is weighted by dividing it by $N_i$, the amount of nodes which have the same predicted label as node $u_i$ and are part of $U'$. 
\begin{table*}[h!]
\centering
\resizebox{\textwidth}{!}{\begin{tabular}{llllllllllrlllr}
\toprule
\multicolumn{1}{l}{\multirow{2}{*}{\% train}} & \multicolumn{1}{c}{\multirow{2}{*}{\textbf{Method}}} & \multicolumn{9}{c}{\textbf{MLDoc}} & & \multicolumn{3}{c}{\textbf{Reuters}}  \\
\cmidrule(l){3-11} \cmidrule(l){12-15}
 & & \textbf{de} & \textbf{en} & \textbf{es}  & \textbf{fr} & \textbf{it} & \textbf{ja} & \textbf{ru} & \textbf{zh} & $\Delta$ & & \textbf{R8}  & \textbf{R52} & $\Delta$ \\
\midrule
\multirow{6}{*}{1\%}
& mBERT & - & - & - & - & - & - & - & - & - &  & 87.4 & 73.4 & 78.6 \\
& XLM-R & - & - & - & - & - & - & - & - & - &  & 87.6 & \textbf{75.0} & \textbf{81.3} \\
& TextGCN & - & - & - & - & - & - & - & - & - &  & 82.8 &65.7 & 74.3 \\ 
& + 2-NR & - & - & - & - & - & - & - & - & - & &  83.9 & 68.5 & 76.2 \\ 
& + Pseudo-label & - & - & - & - & - & - & - & - & - &  & 79.1 & 65.2 & 72.2 \\ 
& FewShotTextGCN & - & - & - & - & - & - & - & - & - & & \textbf{88.6} & 69.2 & 78.9 \\
\midrule
\multirow{6}{*}{2\%} 
& mBERT & 80.2 & 68.0 & 57.1 & 53.8 & 53.2 & 56.6 & 64.0 & 27.1 & 57.5 &  & 83.8 & 78.3 & 81.1 \\
& XLM-R  & 79.4 & 79.0 & 71.1 & 73.5 & 62.4 & 57.5 &\textbf{70.0} & 40.5 & 66.7 &  & 85.8 & \textbf{81.8} & \textbf{83.8} \\
& TextGCN & 60.7 & 68.2 & 71.4 & 35.6 & 59.3 & 63.8 & 55.2 & 73.4 & 61.0 &  & 84.2 & 59.2 & 71.7 \\ 
& + 2-NR & 85.9 & 75.3 & \textbf{75.7} & 81.9 & \textbf{66.9} & 70.7 & 65.7 & 79.1 & 75.2 & &  85.3 & 63.2 & 74.3 \\ 
& + Pseudo-label & 83.7 & 76.2 & 61.0 & 70.0 & 47.1 & 62.6 & 58.1 & 74.8 & 66.9 & & 80.1 & 64.5 & 72.3 \\
& FewShotTextGCN  & \textbf{86.9} &\textbf{84.0} & \textbf{75.7} & \textbf{83.5} & \textbf{66.9} & \textbf{78.9} &67.5 & \textbf{80.3} & \textbf{78.0} & & \textbf{87.2} & 65.2 & 76.2 \\ 
\midrule
\multirow{6}{*}{5\%} 
& mBERT & 89.1 & 85.0 & 74.0 & 84.3 & 67.7 & 77.1 & 73.5 & 80.9 & 79.0  & &  94.7 & 86.0 & 90.4 \\
& XLM-R  & 91.2 & 84.6 & 76.3 & 87.7 & \textbf{75.8} & 81.2 & \textbf{75.2} & \textbf{85.0} & \textbf{82.1} &  & \textbf{95.7} & \textbf{88.8} & \textbf{92.3} \\
& TextGCN & 88.9 & 73.8 & 77.2 & 84.9 & 70.5 & 79.6 & 59.0 & 80.4 & 76.8 & & 87.2& 67.3 & 77.2 \\ 
&  + 2-NR & 89.6 & 85.5 & 79.4 & 86.4 & 75.0 & 81.0 & 67.3 & 81.6 & 80.1 & & 90.4 & 69.0 & 79.7 \\ 
& + Pseudo-label & 88.9 & 87.2 & 77.0 & 83.8 & 72.1 & 79.8 & 60.6 & 80.6 & 78.0 & & 87.4 & 64.4 & 75.9 \\ 
& FewShotTextGCN & \textbf{91.5} & \textbf{88.6} & \textbf{81.2} & \textbf{88.7} & 72.6 & \textbf{81.9} & 70.0 & 82.0 & \textbf{82.1} & & 90.9 & 69.2 & 80.1 \\ 
\midrule
\multirow{6}{*}{10\%} 
& mBERT & 90.3 & 87.5 & 86.3 & 87.3 & 77.1 & 81.2 & 82.5 & 82.8 & 84.4  & &  95.7 & 85.1 & 90.4 \\
& XLM-R  & 91.0 & 88.1 & \textbf{88.8} & 87.2 & 75.8 & 81.9 & \textbf{83.0} & \textbf{88.9} & \textbf{85.6}  & & \textbf{96.9} & \textbf{92.4} & \textbf{94.7} \\
& TextGCN & 90.7 & 85.5 & 87.2 & 86.4 & 72.5 & 81.1 & 72.7 & 85.1 & 82.7 & & 89.1 & 73.6 & 81.4 \\ 
& + 2-NR & 89.9 & 86.8 & 87.8 & 87.1 & \textbf{77.3} & 81.1 & 68.1 & 84.7 & 82.9 & & 90.9 & 76.0 & 83.5 \\ 
& + Pseudo-label & 91.5 & 87.2 & 87.2 & 87.8 & 72.6 & \textbf{82.5} & 74.4 & 85.3 & 83.6 & & 90.2 & 75.6 & 82.9 \\ 
& FewShotTextGCN & \textbf{91.8} & \textbf{90.0} & 87.9 & \textbf{88.7} & 74.4 & \textbf{82.5} & 74.4 & 85.3 & 84.4 & &  92.5 & 80.2 & 86.4 \\ 
\midrule
\multirow{6}{*}{90\%} 
& mBERT & 91.4 & 93.2 & 93.3 & 94.2 & 86.6 & 87.8 & 86.7 & 90.9 & 90.5  & & 95.8 & 94.5 & 95.2 \\
& XLM-R & \textbf{95.2} & \textbf{94.2} & \textbf{95.9} & 93.4 & \textbf{87.1} & 86.9 &\textbf{88.7} & \textbf{91.3} & \textbf{91.4}  & & \textbf{97.5} & \textbf{95.4} & \textbf{96.5} \\
& TextGCN & 94.5 & 91.9 & 94.3 & \textbf{93.6} & 85.8 & \textbf{89.1} & 82.8 & 89.5 & 90.2 & & 94.1 & 82.0 & 88.1 \\ 
& + 2-NR & 94.0 & 91.1 & 93.9 & 92.1 & 84.7 & 86.8 & 84.5 & 88.6 & 89.5 & & 95.3 & 83.1 & 89.2 \\ 
& + Pseudo-label & 94.3 & 92.1 & 94.3 & 92.3 & 85.6 & 88.5 & 82.9 & 89.4 & 89.9 & & 94.1 & 82.3 & 88.2 \\
& FewShotTextGCN & 94.2 & 91.8 & 94.5 & 92.1 & 84.3 & 87.7 & 84.5 & 89.3 & 89.8 & & 95.4 & 85.1 & 90.3 \\ 
\bottomrule                       
\end{tabular}}
\caption{{Average accuracy across 5 different seeds on the test set using a different number of training samples available.} $\Delta$ corresponds to the average accuracy across all datasets/languages. Methods starting with ``+'' correspond to TextGCN extended with one of our corresponding proposed methods at a time. FewShotTextGCN refers to the combination of all our proposed improvements together (i.e., including \textit{both}  2-NR and adaptive pseudo-labeling) as well as our adjusted graph construction method. Highest scoring method per benchmark is marked in \textbf{bold}.} 
\label{tab:low_resource_results}
\end{table*}

\begin{table*}[h!]
\centering
\begin{tabular}{llccccccccr}
\toprule
\multicolumn{1}{l}{\multirow{2}{*}{\% train}} & \multicolumn{1}{c}{\multirow{2}{*}{\textbf{Edge types}}} & \multicolumn{8}{c}{\textbf{MLDoc}} &  \\
\cmidrule(l){3-10}
&  &\textbf{de} & \textbf{en} & \textbf{es}  & \textbf{fr} & \textbf{it} & \textbf{ja} & \textbf{ru} & \textbf{zh} & $\Delta$ \\
\midrule
\multirow{2}{*}{2\%} 
& $\mathcal{V}_d-\mathcal{V}_w$+$\mathcal{V}_w-\mathcal{V}_w$ & 60.7 & 68.2 & 71.3 & 35.6 & 59.3 & 63.8 & 55.2  & 73.4 & 60.9  \\ 
& $\mathcal{V}_d-\mathcal{V}_w$  & \textbf{74.6} &\textbf{ 71.7} & 71.3 & \textbf{76.7} & \textbf{59.9} & \textbf{65.1} & \textbf{59.9} & \textbf{75.4} & \textbf{69.3}  \\ 
\midrule
\multirow{2}{*}{5\%} 
& $\mathcal{V}_d-\mathcal{V}_w$+$\mathcal{V}_w-\mathcal{V}_w$ & 88.4 & 73.7 & 77.2 & \textbf{84.9} & 70.5 & \textbf{79.6} & 59.0  & \textbf{80.4} & 76.7 \\ 
& $\mathcal{V}_d-\mathcal{V}_w$  & \textbf{88.5} & \textbf{80.7} & \textbf{78.1} & 82.7 &\textbf{71.5} & 78.4 & \textbf{60.2} & 80.0 & \textbf{77.5}  \\ 
\midrule
\multirow{2}{*}{10\%} 
& $\mathcal{V}_d-\mathcal{V}_w$+$\mathcal{V}_w-\mathcal{V}_w$  &\textbf{90.7} & 85.5 & \textbf{87.2} & 86.4 & 72.4 & 81.0 & 72.7 & 83.7 & 82.5 \\ 
& $\mathcal{V}_d-\mathcal{V}_w$  & 90.3 & \textbf{86.8} & 87.0 & 86.4 & \textbf{75.8} & \textbf{82.3} & \textbf{74.7} & \textbf{85.1} & \textbf{83.6}  \\ 
\midrule
\multirow{2}{*}{90\%} 
& $\mathcal{V}_d-\mathcal{V}_w$+$\mathcal{V}_w-\mathcal{V}_w$ & \textbf{94.5} & 91.9 & 94.2 & \textbf{93.4} & 85.8 & \textbf{89.1} & 82.8 & \textbf{89.5} & 90.2  \\ 
& $\mathcal{V}_d-\mathcal{V}_w$  & 94.1 & 91.9 & \textbf{94.4} & 93.0 & \textbf{86.6} & 88.7 & \textbf{85.0} & 89.4 & \textbf{90.4} \\ 
\bottomrule
\multirow{2}{*}{\textbf{\#edges}}
& $\mathcal{V}_d-\mathcal{V}_w$+$\mathcal{V}_w-\mathcal{V}_w$ & 7.4M & 11M & 5.5M & 8.4M & 5.2M & 4.9M & 10.2M & 5.2M & 7.2M  \\
& $\mathcal{V}_d-\mathcal{V}_w$ & 1M & 1.3M & 900K & 1.1M & 758K & 1.1M & 1.1M & 889K & 1M \\
\bottomrule                       
\end{tabular}
\caption{{Average accuracy of 5 different seeds on the test set, with a different number of training samples available.} Here, the original TextGCN model is used and only the graph-construction method is varied. $\Delta$ corresponds to the average accuracy across seeds. Highest scoring method per language is marked in \textbf{bold}.} 
\label{tab:results_word_word}
\end{table*}

\section{Experimental setting}

Throughout our experiments, TextGCN is used as a directly comparable baseline. 
Since our main goal is to develop a method that performs well in low-resource scenarios for many languages without the need of any knowledge of that language -- apart from the ability to identify word boundaries in a sentence -- our setup deviates from the original TextGCN work. 
 Unlike the original work, we do not perform a grid-search of hyperparameter settings per experiment/language, but rather keep them fixed -- which make our results not directly comparable to the original.
Similarly to the original TextGCN work, we also consider the R8 and R52 datasets for an analysis of our approach on English (see Section \ref{sec:data}). Additionally, we also provide results for two PLMs trained with the same amount of data. These results are not directly comparable, since the PLMs are trained in an inductive setting, but are included to provide better insight into the positioning of our method in the context of broader literature.
 
 \paragraph{PLM baselines}
 We introduce both multilingual BERT (mBERT) \citep{devlin2018bert} and XLM-R \citep{conneau2019unsupervised} as strong baselines based on the Transformer \citep{vaswani2017attention} architecture. These baselines are fine-tuned in the same data settings, with their architecture settings kept as their defaults as defined in the HuggingFace Transformers library \citep{wolf-etal-2020-transformers}. For training, a learning rate of 5e-5 and a batch-size of 20 is used.
 
\paragraph{Learning settings}
We investigate the effectiveness of our approach when learning from 1\%, 2\%, 5\%, 10\% and 90\% of the available training samples. The 1\% setting is only considered for the R8 and R52 datasets, due to the already relatively small training set size in the MLDoc datasets.
For all settings except the 90\% one, the size of the validation set is equal to the size of the training set (see Appendix \ref{sec:appendix_train_vs_val} for a background experiment on the influence of the division of a limited set of labeled samples over the train and validation sets). The remaining documents are then added to the word--document graph as unlabeled nodes. For the high-resource setting (90\%), the remaining 10\% of the training set is used for validation (i.e., no unlabeled nodes). 
We train all GNN models from scratch for each language and do not rely on any form of transfer- or multi-task learning. 

\paragraph{Training setup and hyperparameters}
We use the Ranger optimizer \cite{liu2019variance,zhang2019lookahead,yong2020gradient}, an adapted version of Adam \cite{kingma2014adam}.
All experiments run for 1000 epochs and the model with the lowest validation loss is used at test time. A learning rate of 0.01 and dropout of 0.5 are used throughout all experiments except when mentioned otherwise. All hidden dimensions are set to 64 and in line with the original TextGCN work, we use two layers of GCN followed by one linear layer for classification. The log schedule for training signal annealing as per Appendix A.2 in \citet{xie2020unsupervised} is used whenever 2-NR is applied. For pseudo-labeling, we set the confidence threshold $\beta = 0.75$ following the original paper.

\section{Results} \label{sec:results}


\subsection{Comparison to TextGCN}
\paragraph{MLDoc}
Table \ref{tab:low_resource_results} shows the results of our experiments. In the 2\% training data setting, FewShotTextGCN outperforms TextGCN by 17\% points on average ($\Delta$) on the eight languages of the MLDoc dataset, showing that we can effectively utilize the properties of heterogeneous word--document graphs to improve learning in low-resource settings in many languages. For MLDoc, which is a dataset with uniform class priors, we see the difference in performance between original TextGCN and TextGCN combined with 2-NR grows larger as the amount of training data decreases, demonstrating that our proposed 2-NR regularizer helps to combat overfitting. Comparing the `+2-NR' results to those of FewShotTextGCN (that uses both 2-NR and adaptive pseudo-labeling), it can be seen that, overall, our regularizer is the primary contributor in outperforming the TextGCN baseline. 
Our version of adaptive pseudo-labeling also outperforms the TextGCN baseline, with the largest margins in the low-resource settings, indicating the effectiveness of utilizing unlabeled document nodes in the word--document graph.

 \noindent In the high-resource (90\%) setting of MLDoc, FewShotTextGCN performs on a par with the original TextGCN. This can be explained by the fact that  2-NR is a regularization method
 and
 the training data set is relatively large in the high-resource setting, which makes that adding regularization to the learning process can be redundant. Furthermore, our version of adaptive pseudo-labeling works based on a set of unlabeled documents not belonging to either the train or the test set, which is a relatively small set of documents in this setting, namely only 10\% of the documents of the total training set. 
 
\paragraph{Reuters}
 Interestingly, FewShotTextGCN outperforms TextGCN consistently in all data settings for the English Reuters datasets, which are highly skewed in their class distribution. This can be seen as supporting evidence for the hypothesis that 2-NR forces the learned feature representations of documents to contain information of all words it contains, which helps to learn distinguishing features for documents of minority classes. 
 
 \subsection{Comparison to PLMs}
\paragraph{MLDoc}
Although FewShotTextGCN only uses $\approx1\%$ of the parameters, has no pretrained knowledge of the considered languages, has no notion of word order in the documents and does not make use of a shared subword vocabulary, it performs on par with large PLMs across all settings for MLDoc. In the lowest resource setting, FewShotTextGCN outperforms all considered PLMs, whereas both PLMs start performing on par as the amount of available data increases. We hypothesize that the somewhat larger difference in performance for the Russian language is attributable to the fact that Russian is a highly inflective language, resulting in many unique words to learn a representation for. The PLMs have the advantage of using a subword vocabulary which serves as a remedy for the formerly described sparsity challenge.

\paragraph{Reuters}
For R8 holds that similarly to the results on MLDoc, FewShotTextGCN outperforms the PLM baselines in the two lowest-resource settings, whereas the PLMs perform better when more training data is available. The results on R52 are more notable, as the gap in performance between FewShotTextGCN and the PLMs grows relatively larger with more available training data. We hypothesize this could be due to the fact for FewShotTextGCN we use only a 64 dimensional hidden size to encode the 52 classes of the dataset, whereas the PLMs use a hidden size of 768.

\section{Ablation experiments}
\label{sec:word-word-edges}
The original TextGCN implementation proposes to use edges between words based on their respective PMI. Since PMI is calculated using a window size of 20, many extra edges are introduced.
For the MLDoc dataset, omitting word--word edges results in a graph that has, on average, only 15\% of the amount of edges compared to the original graph (see Table \ref{tab:results_word_word} for statistics on the number of edges per graph construction method). To analyse the effect of word--word edges, we evaluate the original TextGCN method across the different graph construction methods in the same data availability settings as our main set of experiments (Table \ref{tab:results_word_word}). 
The results provide empirical evidence that, on average, word--word edges are redundant in topic classification problems. The average performance using graphs without word--word edges is always higher; however, performance difference between the two graph construction methods does get smaller as more data is added.
In Appendix \ref{sec:appendix-word-word} we present a visual walk-through of how words can still influence each other's feature representations in a graph without word--word edges. 

\section{Discussion}
\paragraph{K-NR for K > 2}
Here, we argue by example that K-NR can also be applied to other heterogeneous graphs with two or more different kinds of nodes. Consider a network with three kinds of nodes: venue nodes, paper nodes and author nodes \citep{shi2016survey}. Venue nodes have a connection to a paper node if the paper is published at that venue and authors have a connection to the paper node when they are a contributor to that respective paper. No other edges exist on this graph and the classification task concerns the author nodes. In this case, we could apply K-NR on the author nodes based on the intuition that authors that publish a paper at the same venue are more similar to each other than authors that do not publish at the same venue. In order to get from the anchor author node to a positive author node, one has to traverse the graph by hopping to a neighboring paper, venue, paper and finally author node in that respective order -- resulting in $K=4$. On this same graph, K-NR can be applied for paper nodes as well for $K=2$ and traversing via the venue node.
\noindent In general, considering a graph with $M$ different node types, K-NR can be applied if in terms of node types a symmetrical path with an odd number of nodes can be traversed. In this case, $K=2(M-1)$.

\section{Conclusion}
We introduced K-hop Neighborhood Regularization (K-NR), a contrastive learning method for heterogeneous graphs, and showed its implementation for word--document graphs (2-NR) is highly effective in improving learning from scratch in low-resource settings for a range of languages on topic classification tasks. We also showed that we can exploit properties of word--document graphs for improved learning in few-shot settings. We demonstrated that by simplifying the graph construction method via omitting word--word edges we can improve performance while reducing memory requirements in terms of total number of edges.  
Additionally, we showed how pseudo-labeling can be successfully applied to word--document graphs. All approaches combined together form part of our new proposed method, FewShotTextGCN, an improvement over TextGCN for few-shot graph learning. 
FewShotTextGCN performs on par with large PLMs across the considered benchmarks using only a fraction of the parameters and no pretraining whatsoever, showing that GNNs are an attractive alternative for these Transformer-based models.
Finally, using this method, we showed that transductive document classification can be performed successfully on a wide range of typologically diverse languages without any language model pretraining. In future work, we plan to explore the effectiveness of 2-NR on a large range of graphs, such as social networks, citation networks and product--user networks as well as adaptations of K-NR for $K>2$. 

\section{Limitations}
Our work focused on a subset of the text-classification field, namely topic classification. In order to generalize our contributions to other subsets such as sentiment classification, our method might benefit from incorporating word order \citep{johnson2014effective}. Secondly, adding 2-NR to the training process does  slow down the convergence rate of training. Exemplified: regular TextGCN would often reach its lowest validation loss in the range of 50 to 200 update steps, whereas TextGCN + 2-NR would often reach its lowest validation loss in the range of 700 to 900 update steps. We do not consider this a major limitation as all experiments can still be performed on a single GPU with 8Gb of RAM.

\bibliography{anthology,custom}

\begin{thebibliography}{36}
\expandafter\ifx\csname natexlab\endcsname\relax\def\natexlab#1{#1}\fi

\bibitem[{Balntas et~al.(2016)Balntas, Riba, Ponsa, and
  Mikolajczyk}]{balntas2016learning}
Vassileios Balntas, Edgar Riba, Daniel Ponsa, and Krystian Mikolajczyk. 2016.
\newblock Learning local feature descriptors with triplets and shallow
  convolutional neural networks.
\newblock In \emph{Bmvc}, volume~1, page~3.

\bibitem[{Chau and Smith(2021)}]{chau2021specializing}
Ethan~C Chau and Noah~A Smith. 2021.
\newblock Specializing multilingual language models: An empirical study.
\newblock \emph{arXiv preprint arXiv:2106.09063}.

\bibitem[{Chen et~al.(2021)Chen, Ravichandran, and Stolcke}]{chen2021graph}
Long Chen, Venkatesh Ravichandran, and Andreas Stolcke. 2021.
\newblock Graph-based label propagation for semi-supervised speaker
  identification.
\newblock \emph{arXiv preprint arXiv:2106.08207}.

\bibitem[{Conneau et~al.(2019)Conneau, Khandelwal, Goyal, Chaudhary, Wenzek,
  Guzm{\'a}n, Grave, Ott, Zettlemoyer, and Stoyanov}]{conneau2019unsupervised}
Alexis Conneau, Kartikay Khandelwal, Naman Goyal, Vishrav Chaudhary, Guillaume
  Wenzek, Francisco Guzm{\'a}n, Edouard Grave, Myle Ott, Luke Zettlemoyer, and
  Veselin Stoyanov. 2019.
\newblock Unsupervised cross-lingual representation learning at scale.
\newblock \emph{arXiv preprint arXiv:1911.02116}.

\bibitem[{Devlin et~al.(2018)Devlin, Chang, Lee, and
  Toutanova}]{devlin2018bert}
Jacob Devlin, Ming-Wei Chang, Kenton Lee, and Kristina Toutanova. 2018.
\newblock Bert: Pre-training of deep bidirectional transformers for language
  understanding.
\newblock \emph{arXiv preprint arXiv:1810.04805}.

\bibitem[{Ding et~al.(2020)Ding, Wang, Li, Li, and Liu}]{ding2020more}
Kaize Ding, Jianling Wang, Jundong Li, Dingcheng Li, and Huan Liu. 2020.
\newblock Be more with less: Hypergraph attention networks for inductive text
  classification.
\newblock \emph{arXiv preprint arXiv:2011.00387}.

\bibitem[{Grover and Leskovec(2016)}]{grover2016node2vec}
Aditya Grover and Jure Leskovec. 2016.
\newblock node2vec: Scalable feature learning for networks.
\newblock In \emph{Proceedings of the 22nd ACM SIGKDD international conference
  on Knowledge discovery and data mining}, pages 855--864.

\bibitem[{Hamilton et~al.(2017)Hamilton, Ying, and
  Leskovec}]{hamilton2017inductive}
Will Hamilton, Zhitao Ying, and Jure Leskovec. 2017.
\newblock Inductive representation learning on large graphs.
\newblock \emph{Advances in neural information processing systems}, 30.

\bibitem[{Johnson and Zhang(2014)}]{johnson2014effective}
Rie Johnson and Tong Zhang. 2014.
\newblock Effective use of word order for text categorization with
  convolutional neural networks.
\newblock \emph{arXiv preprint arXiv:1412.1058}.

\bibitem[{Kingma and Ba(2014)}]{kingma2014adam}
Diederik~P Kingma and Jimmy Ba. 2014.
\newblock Adam: A method for stochastic optimization.
\newblock \emph{arXiv preprint arXiv:1412.6980}.

\bibitem[{Kipf and Welling(2016)}]{kipf2016semi}
Thomas~N Kipf and Max Welling. 2016.
\newblock Semi-supervised classification with graph convolutional networks.
\newblock \emph{arXiv preprint arXiv:1609.02907}.

\bibitem[{Lee et~al.(2013)}]{lee2013pseudo}
Dong-Hyun Lee et~al. 2013.
\newblock Pseudo-label: The simple and efficient semi-supervised learning
  method for deep neural networks.
\newblock In \emph{Workshop on challenges in representation learning, ICML},
  volume~3, page 896.

\bibitem[{Lewis et~al.(2004)Lewis, Yang, Russell-Rose, and Li}]{lewis2004rcv1}
David~D Lewis, Yiming Yang, Tony Russell-Rose, and Fan Li. 2004.
\newblock Rcv1: A new benchmark collection for text categorization research.
\newblock \emph{Journal of machine learning research}, 5(Apr):361--397.

\bibitem[{Li et~al.(2020)Li, Kumar, Headden, Yin, Wei, Zhang, and
  Yang}]{li2020learn}
Zheng Li, Mukul Kumar, William Headden, Bing Yin, Ying Wei, Yu~Zhang, and Qiang
  Yang. 2020.
\newblock Learn to cross-lingual transfer with meta graph learning across
  heterogeneous languages.
\newblock In \emph{Proceedings of the 2020 Conference on Empirical Methods in
  Natural Language Processing (EMNLP)}, pages 2290--2301.

\bibitem[{Liang et~al.(2020)Liang, Duan, Gong, Wu, Guo, Qi, Gong, Shou, Jiang,
  Cao et~al.}]{liang2020xglue}
Yaobo Liang, Nan Duan, Yeyun Gong, Ning Wu, Fenfei Guo, Weizhen Qi, Ming Gong,
  Linjun Shou, Daxin Jiang, Guihong Cao, et~al. 2020.
\newblock Xglue: A new benchmark datasetfor cross-lingual pre-training,
  understanding and generation.
\newblock In \emph{Proceedings of the 2020 Conference on Empirical Methods in
  Natural Language Processing (EMNLP)}, pages 6008--6018.

\bibitem[{Lin et~al.(2021)Lin, Meng, Sun, Han, Kuang, Li, and
  Wu}]{lin2021bertgcn}
Yuxiao Lin, Yuxian Meng, Xiaofei Sun, Qinghong Han, Kun Kuang, Jiwei Li, and
  Fei Wu. 2021.
\newblock Bertgcn: Transductive text classification by combining gcn and bert.
\newblock \emph{arXiv preprint arXiv:2105.05727}.

\bibitem[{Liu et~al.(2019)Liu, Jiang, He, Chen, Liu, Gao, and
  Han}]{liu2019variance}
Liyuan Liu, Haoming Jiang, Pengcheng He, Weizhu Chen, Xiaodong Liu, Jianfeng
  Gao, and Jiawei Han. 2019.
\newblock On the variance of the adaptive learning rate and beyond.
\newblock \emph{arXiv preprint arXiv:1908.03265}.

\bibitem[{Liu et~al.(2020)Liu, You, Zhang, Wu, and Lv}]{liu2020tensor}
Xien Liu, Xinxin You, Xiao Zhang, Ji~Wu, and Ping Lv. 2020.
\newblock Tensor graph convolutional networks for text classification.
\newblock In \emph{Proceedings of the AAAI conference on artificial
  intelligence}, volume~34, pages 8409--8416.

\bibitem[{McCann(2020)}]{mccann2020fugashi}
Paul McCann. 2020.
\newblock fugashi, a tool for tokenizing japanese in python.
\newblock \emph{arXiv preprint arXiv:2010.06858}.

\bibitem[{McInnes et~al.(2018)McInnes, Healy, and Melville}]{mcinnes2018umap}
Leland McInnes, John Healy, and James Melville. 2018.
\newblock Umap: Uniform manifold approximation and projection for dimension
  reduction.
\newblock \emph{arXiv preprint arXiv:1802.03426}.

\bibitem[{Nikolentzos et~al.(2020)Nikolentzos, Tixier, and
  Vazirgiannis}]{nikolentzos2020message}
Giannis Nikolentzos, Antoine Tixier, and Michalis Vazirgiannis. 2020.
\newblock Message passing attention networks for document understanding.
\newblock In \emph{Proceedings of the AAAI Conference on Artificial
  Intelligence}, volume~34, pages 8544--8551.

\bibitem[{Raffel et~al.(2020)Raffel, Shazeer, Roberts, Lee, Narang, Matena,
  Zhou, Li, Liu et~al.}]{raffel2020exploring}
Colin Raffel, Noam Shazeer, Adam Roberts, Katherine Lee, Sharan Narang, Michael
  Matena, Yanqi Zhou, Wei Li, Peter~J Liu, et~al. 2020.
\newblock Exploring the limits of transfer learning with a unified text-to-text
  transformer.
\newblock \emph{J. Mach. Learn. Res.}, 21(140):1--67.

\bibitem[{Rajpurkar et~al.(2016)Rajpurkar, Zhang, Lopyrev, and
  Liang}]{rajpurkar2016squad}
Pranav Rajpurkar, Jian Zhang, Konstantin Lopyrev, and Percy Liang. 2016.
\newblock Squad: 100,000+ questions for machine comprehension of text.
\newblock \emph{arXiv preprint arXiv:1606.05250}.

\bibitem[{Schwenk and Li(2018)}]{schwenk2018corpus}
Holger Schwenk and Xian Li. 2018.
\newblock A corpus for multilingual document classification in eight languages.
\newblock \emph{arXiv preprint arXiv:1805.09821}.

\bibitem[{Shi et~al.(2016)Shi, Li, Zhang, Sun, and Philip}]{shi2016survey}
Chuan Shi, Yitong Li, Jiawei Zhang, Yizhou Sun, and S~Yu Philip. 2016.
\newblock A survey of heterogeneous information network analysis.
\newblock \emph{IEEE Transactions on Knowledge and Data Engineering},
  29(1):17--37.

\bibitem[{Strubell et~al.(2019)Strubell, Ganesh, and
  McCallum}]{strubell-etal-2019-energy}
Emma Strubell, Ananya Ganesh, and Andrew McCallum. 2019.
\newblock \href {https://doi.org/10.18653/v1/P19-1355} {Energy and policy
  considerations for deep learning in {NLP}}.
\newblock In \emph{Proceedings of the 57th Annual Meeting of the Association
  for Computational Linguistics}, pages 3645--3650, Florence, Italy.
  Association for Computational Linguistics.

\bibitem[{van~der Heijden et~al.(2020)van~der Heijden, Abnar, and
  Shutova}]{van2020comparison}
Niels van~der Heijden, Samira Abnar, and Ekaterina Shutova. 2020.
\newblock A comparison of architectures and pretraining methods for
  contextualized multilingual word embeddings.
\newblock In \emph{Proceedings of the AAAI Conference on Artificial
  Intelligence}, volume~34, pages 9090--9097.

\bibitem[{Vaswani et~al.(2017)Vaswani, Shazeer, Parmar, Uszkoreit, Jones,
  Gomez, Kaiser, and Polosukhin}]{vaswani2017attention}
Ashish Vaswani, Noam Shazeer, Niki Parmar, Jakob Uszkoreit, Llion Jones,
  Aidan~N Gomez, {\L}ukasz Kaiser, and Illia Polosukhin. 2017.
\newblock Attention is all you need.
\newblock In \emph{Advances in neural information processing systems}, pages
  5998--6008.

\bibitem[{Wang et~al.(2019)Wang, Pruksachatkun, Nangia, Singh, Michael, Hill,
  Levy, and Bowman}]{wang2019superglue}
Alex Wang, Yada Pruksachatkun, Nikita Nangia, Amanpreet Singh, Julian Michael,
  Felix Hill, Omer Levy, and Samuel Bowman. 2019.
\newblock Superglue: A stickier benchmark for general-purpose language
  understanding systems.
\newblock \emph{Advances in neural information processing systems}, 32.

\bibitem[{Wang et~al.(2021)Wang, Liu, Yang, Liu, and Wang}]{wang2021cross}
Ziyun Wang, Xuan Liu, Peiji Yang, Shixing Liu, and Zhisheng Wang. 2021.
\newblock Cross-lingual text classification with heterogeneous graph neural
  network.
\newblock \emph{arXiv preprint arXiv:2105.11246}.

\bibitem[{Wolf et~al.(2020)Wolf, Debut, Sanh, Chaumond, Delangue, Moi, Cistac,
  Rault, Louf, Funtowicz, Davison, Shleifer, von Platen, Ma, Jernite, Plu, Xu,
  Scao, Gugger, Drame, Lhoest, and Rush}]{wolf-etal-2020-transformers}
Thomas Wolf, Lysandre Debut, Victor Sanh, Julien Chaumond, Clement Delangue,
  Anthony Moi, Pierric Cistac, Tim Rault, Rémi Louf, Morgan Funtowicz, Joe
  Davison, Sam Shleifer, Patrick von Platen, Clara Ma, Yacine Jernite, Julien
  Plu, Canwen Xu, Teven~Le Scao, Sylvain Gugger, Mariama Drame, Quentin Lhoest,
  and Alexander~M. Rush. 2020.
\newblock \href {https://www.aclweb.org/anthology/2020.emnlp-demos.6}
  {Transformers: State-of-the-art natural language processing}.
\newblock In \emph{Proceedings of the 2020 Conference on Empirical Methods in
  Natural Language Processing: System Demonstrations}, pages 38--45, Online.
  Association for Computational Linguistics.

\bibitem[{Xie et~al.(2020)Xie, Dai, Hovy, Luong, and Le}]{xie2020unsupervised}
Qizhe Xie, Zihang Dai, Eduard Hovy, Thang Luong, and Quoc Le. 2020.
\newblock Unsupervised data augmentation for consistency training.
\newblock \emph{Advances in Neural Information Processing Systems},
  33:6256--6268.

\bibitem[{Yao et~al.(2019)Yao, Mao, and Luo}]{yao2019graph}
Liang Yao, Chengsheng Mao, and Yuan Luo. 2019.
\newblock Graph convolutional networks for text classification.
\newblock In \emph{Proceedings of the AAAI conference on artificial
  intelligence}, volume~33, pages 7370--7377.

\bibitem[{Yong et~al.(2020)Yong, Huang, Hua, and Zhang}]{yong2020gradient}
Hongwei Yong, Jianqiang Huang, Xiansheng Hua, and Lei Zhang. 2020.
\newblock Gradient centralization: A new optimization technique for deep neural
  networks.
\newblock \emph{arXiv preprint arXiv:2004.01461}.

\bibitem[{Zhang et~al.(2019)Zhang, Lucas, Ba, and Hinton}]{zhang2019lookahead}
Michael Zhang, James Lucas, Jimmy Ba, and Geoffrey~E Hinton. 2019.
\newblock Lookahead optimizer: k steps forward, 1 step back.
\newblock In \emph{Advances in Neural Information Processing Systems}, pages
  9597--9608.

\bibitem[{Zhou et~al.(2019)Zhou, Shi, Zhang, Huang, and Li}]{zhou2019effective}
Ziang Zhou, Jieming Shi, Shengzhong Zhang, Zengfeng Huang, and Qing Li. 2019.
\newblock Effective semi-supervised node classification on few-labeled graph
  data.
\newblock \emph{arXiv preprint arXiv:1910.02684}.

\end{thebibliography}
\bibliographystyle{acl_natbib}

\appendix
\section{2-NR intuition and visualization} \label{sec:appendix_2nr}
\begin{figure*}[ht]
    \centering
    \includegraphics[width=\textwidth]{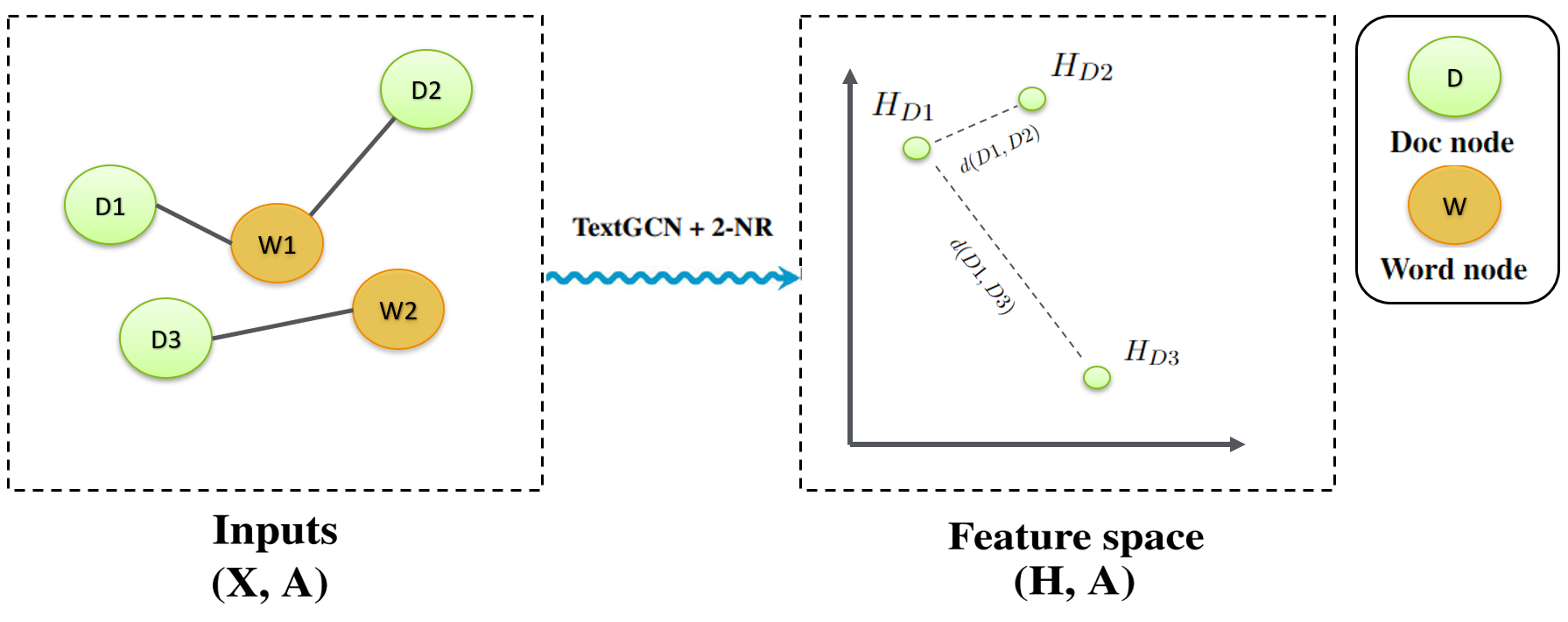}
    \caption{\textbf{2-hop Neighborhood Regularization visualized for a simple graph.} On the left, the graph is defined as a set of nodes $X$ and an adjacency matrix $A$. The right figure depicts the transformed features $H$ of each document $D_i$. Documents $D1$ and $D2$ have word $W1$ as shared neighbor and hence their distance in feature space (depicted by the dotted line with $d(D1, D2)$) gets smaller, whereas the feature representations of $D1$ and $D3$ get pushed away from each other, resulting in a larger distance in the respective space ($d(D1, D3)$). Word nodes are omitted in feature space (right-hand side) for demonstration purposes.}
    \label{fig:2-NR}
\end{figure*}

In Figure \ref{fig:2-NR}, we present a visualization of the process behind 2-NR. 
Intuitively, 2-NR forces the model to incorporate information of all words contained in it such that documents with shared neighbors (i.e., overlap in vocabulary) are closer to each other in semantic space than documents without shared neighbors in feature space, resulting in learning better feature representations of documents.  

\section{Dataset statistics}
Table \ref{tab:statistics} presents an overview of data statistics.
{\small
\begin{table*}[t]\footnotesize
\centering
\renewcommand{\arraystretch}{1.2}
\caption{Summary statistics of datasets.}
\begin{tabular}{c|ccccccc}
\hline
\bf{Dataset}& \bf{\# Docs}	& \bf{\# Training}& \bf{\# Test}& \bf{\# Words} & \bf{\# Nodes}& \bf{\# Classes} & \bf{Average Length} \\
\hline
MLDoc-de & 5,000 & 1,000 & 4,000 & 14,358 & 19,358 & 4 & 144 \\
MLDoc-en & 5,000 & 1,000 & 4,000 & 17,665 & 22,665 & 4 & 215.6 \\
MLDoc-es & 5,000 & 1,000 & 4,000 & 11,662 & 16,662 & 4 & 143.2 \\
MLDoc-fr & 5,000 & 1,000 & 4,000 & 15,231 & 20,231 & 4 & 175.8 \\
MLDoc-it & 5,000 & 1,000 & 4,000 & 10,075 & 15,075 & 4 & 103.8 \\
MLDoc-ja & 5,000 & 1,000 & 4,000 & 8,423 & 13,423 & 4 & 271 \\
MLDoc-ru & 5,000 & 1,000 & 4,000 & 19,786 & 24,786 & 4 & 167.3 \\
MLDoc-zn & 5,000 & 1,000 & 4,000 & 9,270 & 14,270 & 4 & 163.2 \\
 R8 & 7,674 &5,485 & 2,189 & 7822 & 15,496 & 8 & 98.9\\ 
 R52& 9,100 & 6,532	 & 2,568 & 9027 & 17,992 & 52 & 106.3 \\
\hline
\end{tabular}
\label{tab:statistics}
\end{table*}
    }

\section{Visualization of document embeddings} \label{sec:appendix_umap_comparison}
\begin{figure*}
\centering
\begin{subfigure}{.5\textwidth}
  \centering
  \includegraphics[width=\linewidth]{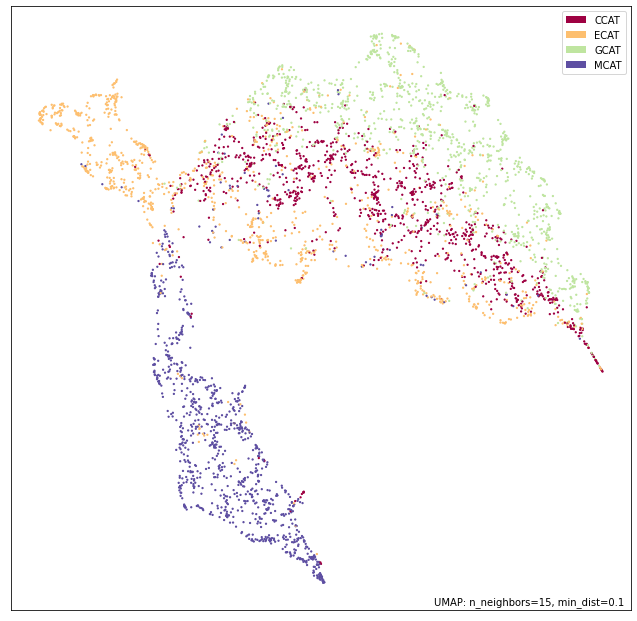}
  \caption{Original TextGCN}
  \label{fig:sub1}
\end{subfigure}%
\begin{subfigure}{.5\textwidth}
  \centering
  \includegraphics[width=\linewidth]{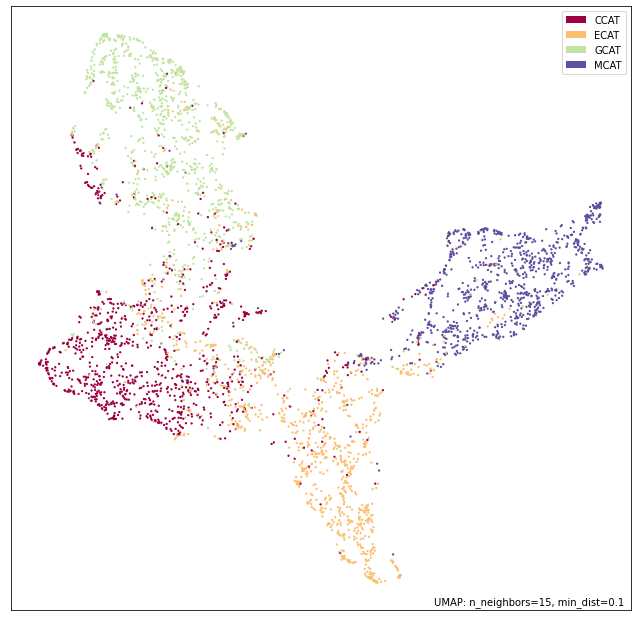}
  \caption{FewShotTextGCN}
  \label{fig:sub2}
\end{subfigure}
\caption{Visualization of the test set document embeddings in the final feature space of TextGCN and FewShotTextGCN on MLDoc-de using the 2\% training data setting. UMAP \citep{mcinnes2018umap} is used for dimensionality reduction}
\label{fig:umap_viz}
\end{figure*}

Figure \ref{fig:umap_viz} shows an example of the difference in class separability between TextGCN and our method, FewShotTextGCN. It can be easily seen that using FewShotTextGCN there is less overlap between regions in which instances of the respective classes live, which is in line with the observations in Table \ref{tab:low_resource_results} where FewShotTextGCN outperforms TextGCN by 26\% points absolute accuracy.

\section{Effective use of data on a limited budget: training vs validation} \label{sec:appendix_train_vs_val}
\begin{figure*}
    \centering
    \includegraphics[width=\textwidth]{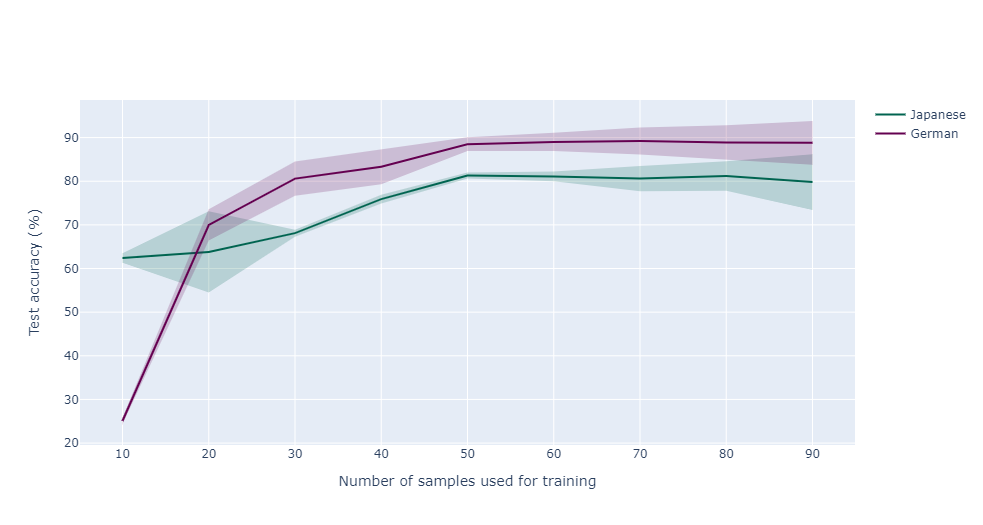}
    \caption{Average test accuracy and standard deviation for 10 seeds with different amounts of data used for training. Given a total of 100 labeled samples, we vary how many are used for training and validation, adjusted in steps of 10.}
    \label{fig:train_vs_val}
\end{figure*}
Figure \ref{fig:train_vs_val} shows the average accuracy and standard deviation of 10 seeds when learning from a total of 100 labeled samples, divided over the train and validation set for Japanese and German, using the original TextGCN model. Other languages are omitted from the plot to prevent visual cluttering, but follow a similar trend: when using too little data for training, the model fails to learn to generalize well -- as can be seen by the relatively low average accuracy on the test set. For German, it is even the case that learning does not converge at all when training on 10 samples, as we observe a mean accuracy of 25\% when the number of training samples is 10. The balance between achieving the highest accuracy with the smallest standard deviation seems to be around the 50/50 split point. Increasing the training data at the cost of fewer validation data after this point can, in some cases, such as for Japanese at the 80/20 point, result in slightly higher accuracy, but the standard deviation across different seeds also increases, confirming the importance of a good-sized validation set during learning. 

\section{On the usefulness of word--word edges in word--document graphs: illustration} \label{sec:appendix-word-word}
\begin{figure*}
\centering
\begin{subfigure}[t]{.45\textwidth}
  \centering
  \includegraphics[width=\linewidth]{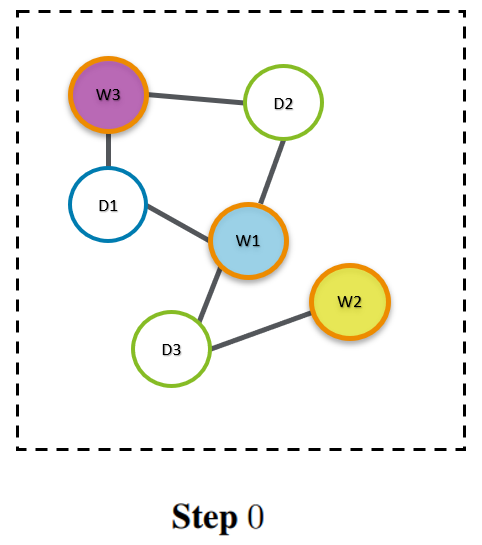}
  \caption{\textbf{Words influencing each other's feature representation without word--word edges over the course of training: initialization.} Classes are represented in different node outline color (i.e., two document classes \textit{green, blue}) and words with an \textit{orange} node outline. The actual color of nodes represents the semantic value encoded in the initial embedding. We simplify the graph by omitting the initial features of the document nodes $D1$, $D2$ and $D3$ and assume they will be learned based on the features of the words occurring in them. Since both $W1$ and $W3$ occur in $D1$ and $D2$, they can intuitively not be distinguishing features for either the \textit{green} or the \textit{blue} class.}
  \label{fig:word-word-init}
\end{subfigure}%
\hspace{2em}
\begin{subfigure}[t]{.45\textwidth}
  \centering
  \includegraphics[width=\linewidth]{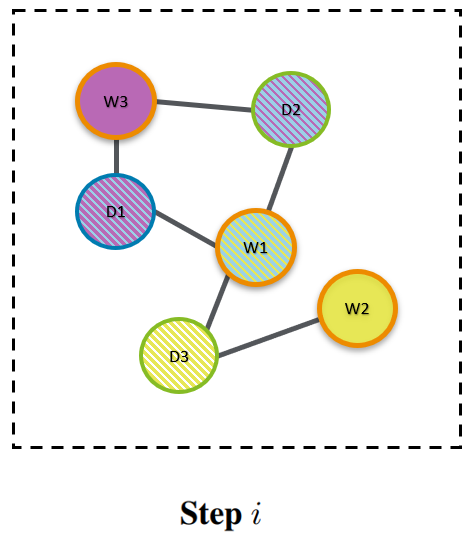}
  \caption{\textbf{Words influencing each other's feature representation without word--word edges over the course of training: during the course of training.} Classes are represented in different node outline color (i.e., two document classes \textit{green, blue}) and words with an \textit{orange} node outline. The actual color of nodes represents the semantic value encoded in the initial embedding. After $i$ steps of training, $D3$ will have incorporated some of the information of $W2$ in its own embedding (depicted by the yellow stripes in the node) in order for it to be a good predictor of the \textit{green} class. Similarly, $D1$ will have incorporated features of $W3$ in its own embedding (depicted by the purple stripes in the node), as it is the most distinguishing word for the \textit{blue} class. Finally, if we consider two-layer GNNs, the feature representation at layer 1 of $W1$, denoted by $H_{W1}^1$, is a function of its own embedding and the aggregated messages from $H_{D1}^0$ and $H_{D2}^0$ and $H_{D3}^0$. Similarly, the final feature representation of $D2$, $H_{D2}^2$, is a function of $H_{D1}^1$ and aggregated messages from $H_{W1}^1$ and $H_{W2}^1$. In order to perform the right classification, $H_{D2}^2$, the information corresponding to the \textit{green} class (yellow filling of node) should be passed on from $H_{D3}^0$, which is the document embedding of $D3$, to $H_{W1}^1$, and then finally to $H_{D2}^2$. Due to the nature of GCNs, no distinction can be made between what message is sent from what node -- as per Equation \ref{eq:gcn}, $x_i = H_{W1}^0$, it can be easily seen that optimizing $H_{W1}^0$ to be a good predictor for the \textit{green} class (i.e., adjusting its feature representation to the yellow color), helps in classifying $D2$ correctly.}
  \label{fig:sub2}
\end{subfigure}
\caption{A visualization with guiding explanation to provide an intuition behind how features of words might still influence each other with having direct edges between them}
\label{fig:word-word-progress}
\end{figure*}

We argue that the added benefit of word--word edges is limited based on the premises that: 1) words occurring in the same window might not influence each other's meaning at all, especially since sentence boundaries are not taken into account; 2) only global information of word co-occurrence (over the whole corpus) is considered, meaning that word A might be connected to word B, but they might not co-occur in document X -- yet, they still influence each other's feature representation as much as when they would have co-occurred in that document; 3) over the course of training, words co-occurring in a document can still influence each other's representation, under the assumption that at least a two-layer GNN is used. We illustrate this by visualizing the initial node embeddings over the course of a hypothetical training schedule, starting at a random initialization in Figure \ref{fig:word-word-init} and going up until some training step $i$ in Figure \ref{fig:word-word-progress}, where the model has improved its performance in the classification task.

\end{document}